\documentclass[sigplan]{acmart}

\usepackage[T1]{fontenc}    
\usepackage{hyperref}       
\usepackage{url}            
\usepackage{booktabs}       
\usepackage{amsfonts}       
\usepackage{nicefrac}     
\usepackage{xcolor}
\usepackage{microtype}      
\usepackage{verbatim}
\usepackage[linesnumbered,ruled,vlined]{algorithm2e}
\SetKwInput{KwInput}{Require}                
\SetKwInput{KwOutput}{Ensure} 
\usepackage{amsthm}
\usepackage{graphicx}
\usepackage[skip=2pt]{subcaption}
\usepackage{setspace}
\usepackage{amsmath}
\usepackage{graphics}

\theoremstyle{definition}
\newtheorem{definition}{Definition}
\newtheorem{lemma}{Lemma}

\SetCommentSty{mycommfont}

\makeatletter
\renewcommand\@formatdoi[1]{\ignorespaces}
\makeatother
\renewcommand\footnotetextcopyrightpermission[1]{} 
\AtBeginDocument{%
  \providecommand\BibTeX{{%
    \normalfont B\kern-0.5em{\scshape i\kern-0.25em b}\kern-0.8em\TeX}}}

\setcopyright{acmcopyright}
\copyrightyear{2018}
\acmYear{2018}
\acmDOI{10.1145/1122445.1122456}

\begin{document}

\title{DPD-InfoGAN: Differentially Private Distributed InfoGAN}

\author{Vaikkunth Mugunthan}
\authornote{Equal Contribution}
\email{vaik@mit.edu}
\affiliation{%
  \institution{Massachusetts Institute of Technology}
  \streetaddress{}
  \city{Cambridge}
  \state{MA}
  \country{USA}
  \postcode{02139}
}

\author{Vignesh Gokul\footnotemark[1]}
\email{vgokul@eng.ucsd.edu}
\affiliation{%
  \institution{University of California, San Diego}
  \streetaddress{}
  \city{San Diego}
  \state{CA}
  \country{USA}}

\author{Lalana Kagal}
\email{lkagal@mit.edu}
\affiliation{%
  \institution{Massachusetts Institute of Technology}
  \streetaddress{}
  \city{Cambridge}
  \state{MA}
  \country{USA}
  \postcode{02139}
}

\author{Shlomo Dubnov}
\email{sdubnov@ucsd.edu}
\affiliation{%
  \institution{University of California, San Diego}
  \streetaddress{}
  \city{San Diego}
  \state{CA}
  \country{USA}}

\renewcommand{\shortauthors}{Trovato and Tobin, et al.}

\begin{abstract}
Generative Adversarial Networks (GANs) are deep learning architectures capable of generating synthetic datasets. Despite producing high-quality synthetic images, the default GAN has no control over the kinds of images it generates. The Information Maximizing GAN (InfoGAN) is a variant of the default GAN that introduces feature-control variables that are automatically learned by the framework, hence providing greater control over the different kinds of images produced. Due to the high model complexity of InfoGAN, the generative distribution tends to be concentrated around the training data points. This is a critical problem as the models may inadvertently expose the sensitive and private information present in the dataset. To address this problem, we propose a differentially private version of InfoGAN (DP-InfoGAN). We also extend our framework to a distributed setting (DPD-InfoGAN) to allow clients to learn different attributes present in other clients' datasets in a privacy-preserving manner. In our experiments, we show that both DP-InfoGAN and DPD-InfoGAN can synthesize high-quality images with flexible control over image attributes while preserving privacy.
\end{abstract}

\ccsdesc[500]{Security and privacy~Domain-specific security and privacy architectures}
\ccsdesc[500]{Computing methodologies~Machine learning}
\ccsdesc[500]{Computing methodologies~Computer vision}
\keywords{InfoGAN, Differential Privacy, Distributed Learning, Deep Learning}

\maketitle
\pagestyle{plain}

\section{Introduction}
\label{sec:intro}

Deep Neural Networks can be used to train high-quality models with state-of-the-art performance in a myriad of applications, including medical image analysis, health informatics, language representation, and many more. However, building such models is not an easy task as it requires access to a large amount of high-quality data. Sharing private data is not an option in many scenarios due to regulations such as GDPR \cite{voigt2017eu}.

GANs \cite{goodfellow2014generative}, a class of Generative models \cite{rezende2014stochastic,li2015generative,makhzani2015adversarial}, can be used to alleviate this arduous data-collection problem. GANs can learn the distribution of training data and generate high-quality fake data samples that have a distribution similar to the original distribution. Ideally, GANs can be used to protect the privacy of individuals in the dataset as they reveal only the distribution and not the sensitive private data of individuals. Despite this property, GANs may potentially expose the private information of training samples as they don't provide guarantees on what information the fake data may reveal about the sensitive training data. Machine learning models, including GAN models, are susceptible to a multitude of attacks including reconstruction and membership inference attacks \cite{shokri2017membership,nasr2018comprehensive,melis2019exploiting,hayes2019logan}, demonstrating that additional privacy is required in the form of protecting model parameters. These attacks can be addressed through the use of differential privacy \cite{dwork2006calibrating}. 

Differential privacy is the state-of-the-art model for protecting the privacy of individuals in a statistical dataset. It ensures that an adversary cannot infer if a particular individual's record is included in the dataset, hence providing the necessary guarantees to train privacy-preserving models on sensitive data. In recent times there have been studies on differentially private GANs \cite{torkzadehmahani2019dp,jordon2018pate, chen2018differentially,frigerio2019differentially,xie2018differentially}. However, most of these methods are focused on generating fixed synthetic data (with or without labels) and do not provide flexibility in controlling attributes of the synthetic data. For example, synthesizing images with different attributes (e.g.\ thickness, rotation, pose, etc.) involve separately training a new model with a new dataset in a private manner, which is expensive. Instead, we leverage InfoGAN \cite{chen2016infogan} to facilitate control over the generated images, while preserving the privacy of the generator.

In this paper, we propose a differentially private framework for InfoGAN and evaluate it on the MNIST dataset \cite{lecun2010mnist}. Our experiments show that our framework can synthesize high-quality images with strong privacy guarantees. We also analyze the trade-off between privacy and quality of control over the generated images. 

Also, we propose a distributed InfoGAN (DPD-InfoGAN) with a shared Q network to capture various attributes of images owned by different clients in a privacy-preserving manner. This allows clients with limited training data to learn intricate features present in the datasets of other clients. For example, if different clients own a subset of MNIST data, then each of the clients would not be exposed to all the variances in the images (e.g. all possible rotation angles, thickness factors, etc) but will still learn to synthesize such characteristics. Aggregating such models using federated learning \cite{konevcny2016federated} would be an expensive process as large model parameters have to be shared and aggregated every round. To overcome this problem, our approach uses a shared Q network in a distributed setting to decrease the number of parameters exchanged and henceforth reducing communication costs.

We show that our paradigm of training distributed InfoGANs enables each client to learn rich and varied feature representations (controlling attributes of generated images) when compared with a single client setting with the same number of images.

\section{Background}

\subsection{InfoGAN} 

Generative Adversarial Networks involve training two networks simultaneously: a discriminator $D$ and a generator $G$. The generator maps a latent space ($p(z)$) to a fake distribution. The discriminator tries to discriminate between real data ($p(x)$) and the fake distribution. The two networks compete with each other in an adversarial setup, i.e., the generator tries to fool the discriminator into classifying its distribution as real data, while the discriminator aims to correctly classify fake and real images. This leads to a minimax game as follows:
\begin{align*}
 \min_{G} \max_{D} V(D,G) = E_{x \sim p_{(x)}}[log(D(x))] + \\ E_{z \sim p(z)}[log(1 - D(G(x))]
\end{align*}
InfoGAN proposes a framework to disentangle the latent space of GANs in an unsupervised manner. The goal is to disentangle the latent space such that meaningful semantics of the data distribution are captured. The input to the generator is split into two components: noise and latent codes (from prior $p(c)$). The latent codes can be discrete or continuous. The latent codes are made meaningful by maximizing the mutual information between the generated data points and the codes. The authors use an auxiliary distribution $Q(c|x)$ to approximate the posterior, modifying the minimax game as follows:
\begin{align}
 \min_{G,Q} \max_{D} V_{InfoGAN}(D,G,Q) = V(D,G) - \lambda L_{I}(G,Q),
\end{align}
 where $L_{I}(G,Q)$ is given by
\begin{align}
    L_{I}(G,Q) = E_{c \sim p(c), x \sim G(z,c)} [log Q(c|x)] + H(c),
\end{align}
where $H(c)$ is the entropy of the prior and is treated as constant, and $\lambda$ is a hyperparameter and is set to 1. We choose p(c) as the Gaussian Distribution with zero mean and unit standard deviation. p(x) refers to the real data distribution and G(z,c) refers to the output distribution of the generator G.       
\subsection{Differential Privacy }
Differential privacy \cite{dwork2006calibrating,dwork2014algorithmic} is a notion of privacy that ensures that statistical analysis does not compromise privacy by requiring that two datasets that are differing by a single individual should be statistically indistinguishable.

\begin{definition}($(\epsilon, \delta)$-Differential Privacy)
     A randomized mechanism $\mathcal{M}$ satisfies $(\epsilon,\delta)$-differential privacy ($(\epsilon,\delta)$-DP) when there exists $\epsilon > 0$, $\delta > 0$, such that 
    \begin{align}
    \text{Pr [}\mathcal{M}(D_1)\in S\text{]}\leq e^\epsilon \text{Pr [}\mathcal{M}(D_2)\in S\text{]} + \delta
     \end{align}
    holds for every $S \subseteq $ Range($\mathcal{M}$) and for all datasets $D_1$ and $D_2$ differing on at most one element. 
\end{definition}
\begin{lemma}\cite{xie2018differentially}
In order to guarantee $(\epsilon, \delta)$-Differential privacy for the discriminator, we assign the following value to the noise scale $\sigma_n$ :
\begin{align}
        \sigma_n= \frac{ 2p\sqrt{I_{d}log(\frac{1}{\delta}) } }{\epsilon },
        \label{sig}
\end{align}

where the sampling probability $p$ =
$\frac{n}{N}$ (n represents the batch size, N represents the dataset size), $I_{d}$ is the number of discriminator iterations for every generator iteration, $\epsilon$ is the privacy-loss parameter, and $\delta$ is the privacy violation parameter.
\end{lemma}

\section{Our Approach}
\vspace{-10pt}
\begin{figure}[h]
\includegraphics[width=8cm]{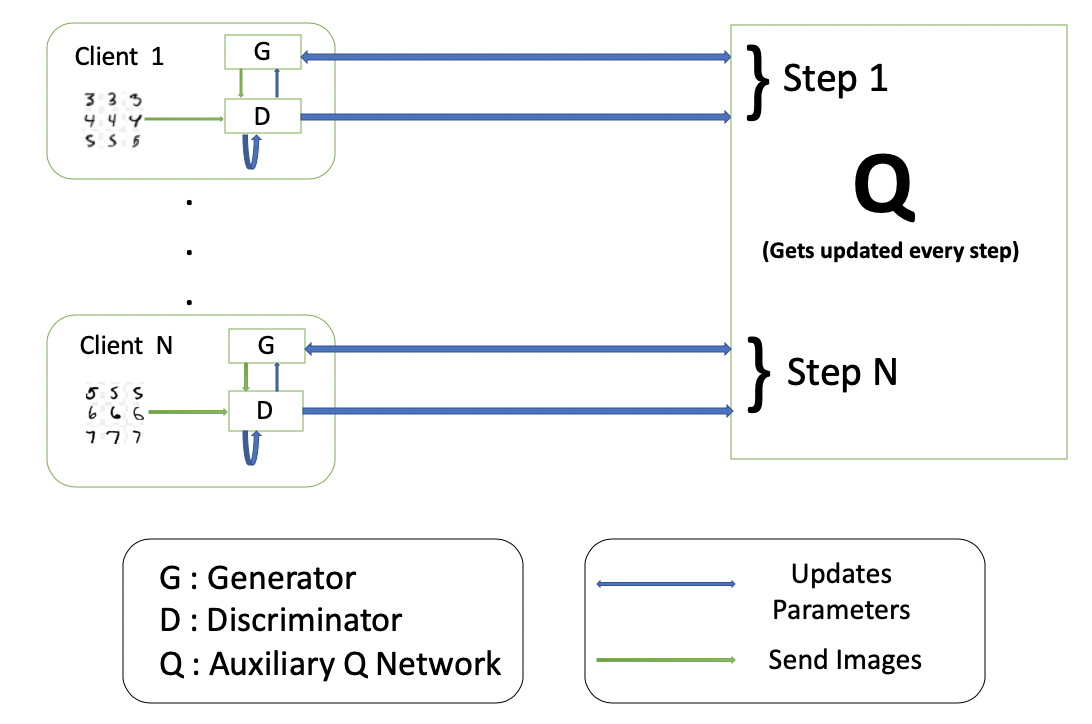}
\caption{Framework for 1 Round}
\label{framework}
\end{figure}

The details of our method to achieve a privacy-preserving InfoGAN are shown in Algorithm \ref{DPInfoGAN}. After computing the gradients of the discriminator (line 4-7), we clip them with clipping parameter $C_p$ (line 8) to bound the gradients. We set $I_d=1$ in Equation \ref{sig} and compute noise scale $\sigma_n$. We add noise to the gradients and then update the discriminator weights using the Adam optimizer \cite{kingma2014adam} (line 10). The $NLL$ in line 12 refers to the Negative Log-Likelihood loss.

The training of the Q network is differentially private due to the post-processing property \cite{dwork2014algorithmic}, as the Q network operates on top of the discriminator. Similarly, the generator satisfies differential privacy, as the generator receives updates from the discriminator and the Q network which are trained in a differentially private manner. We can also keep track of the privacy budget spent in our algorithm by using Moment Accountant \cite{abadi2016deep} or Renyi DP accountant \cite{mironov2017renyi}.

In the distributed setting (Figure \ref{framework}), a similar method is used for N clients, where each client consists of a generator and a discriminator. All the clients make use of a single auxiliary network Q. As explained in Algorithm \ref{DPInfoGAN2}, the Q network is updated sequentially by each client in a given round. That is, in the same round, each client accesses the Q network that had been updated by the previous client. A single Q network is responsible for providing estimates of codes for all the clients. This reduces the communication cost as we only share the outputs of the discriminator (from client to Q-network) and the Q network (from Q-network to client), rather than sending the entire models of the of the generator, discriminator, and Q network as in the case of federated learning. Hence the communication load is massively reduced and therefore cost-efficient than FL.
\begin{algorithm}
\setstretch{1.25}
\DontPrintSemicolon
  
  \KwInput{Clipping parameter for gradients $C_p$, Noise Scale $\sigma_n$, Discriminator $D$, Generator $G$, Auxiliary network providing estimate of the code $Q$, Real data points $X= (x_1,x_2, \dots,x_M)$, Batch size $m$, Noise prior $p(z)$, Latent code prior $p(c)$, Learning Rate $\alpha$}
  \KwOutput{Differentially Private Generator $\theta_g$}
  Sample batch $x = \{x_{i}\}_{i=1}^{m}$ from real data points $X$ \\
  Sample noise $z =\{z_i\}_{i=1}^m$ from noise prior $p(z)$  \\
  Sample codes $c = \{c_i\}_{i=1}^m$ from prior $p(c)$\\
  \tcc{\textbf{Compute batch loss for discriminator}}
  \For{i = 1 to m}
  {
  $D_{loss}(x_i,z_i,c_i) := log (D(x_i)) + D(1 - log(D(G(z_i, c_i))))$\\
  \tcc{\textbf{Calculate gradients with respect to discriminator weights}}
  $grad_d(x_i,z_i,c_i) := \nabla_{\theta_d} D_{loss}(x_i,z_i,c_i) $\\
  }
  \tcc{\textbf{Clip gradients to bound them}}
  $grad_d(x,z,c) := grad_d(x,z,c)/max(1.0, ||grad_d(x,z,c)||_2/C_p) $\\
  \tcc{\textbf{Compute average gradient for batch}}
  $grad_d(x,z,c) = (1/m)*\Sigma_{i=1}^m grad_d(x_i,z_i,c_i)$ \\
 
  \tcc{\textbf{Add noise to make discriminator differentially private}}
  $grad_d(x,z,c) := grad_d(x,z,c) + (1/m)* N(0,\sigma_{n}^2C^2I)$ \\
  \tcc{\textbf{Update weights of the discriminator using Adam optimizer}}
  $\theta_{d_{new}} := \theta_{d} - \alpha.ADAM(grad_d(x,z,c), \theta_d)$ \\
  \tcc{\textbf{Calculate estimate of codes from Q}}
  $Q_{logits}, mean, variance = Q(D(G(z,c)))$\\
  \tcc{\textbf{Compute the Negative Log Likelihood of  target codes and estimate}}
  $Q_{loss} = NLL(c, mean, variance, Q_{logits})$\\
  \tcc{\textbf{Compute loss and gradients for generator}}
  $G_{loss} := D(1 - log(D(G(z, c)))) + Q_{loss}$\\
  $grad_g, grad_q := \nabla_{\theta_g} G_{loss}, \nabla_{\theta_q} G_{loss}$\\ 
  $\theta_{g_{new}} := \theta_{g} - \alpha.ADAM(grad_g, \theta_g)$ \\
  $\theta_{q_{new}} := \theta_{q} - \alpha.ADAM(grad_q, \theta_q)$ \\
  return $\theta_{g_{new}}, \theta_{d_{new}}, \theta_{q_{new}} $
\caption{Differentially Private InfoGAN (DP-InfoGAN)}
\label{DPInfoGAN}
\end{algorithm}

\begin{algorithm}
\setstretch{1.25}
\DontPrintSemicolon
  
  \KwInput{Clients $C = (C_1,C_2, \dots, C_N)$, where $C_i = (G_i,D_i)$, Auxiliary network providing estimate of the code $Q$, Total number of rounds per client $R$}
  \KwOutput{Differentially Private Generator $G_i$}
  \For{$r$ = 1 to $R$}
  {
  \For{$i$ = 1 to $N$}
  {
  \tcc{\textbf{Train $\mathbf{C_i}$ using Algorithm \ref{DPInfoGAN}}}
  $\theta_{g_{new}}, \theta_{d_{new}}, \theta_{q_{new}}$ = Algorithm1($G_i$, $D_i$, $Q$)
  \tcc{\textbf{Update Q weights}}
  $Q$.weights = $\theta_{q_{new}}$
  
  }
  }
  return $C$, $Q$
\caption{Differentially Private Distributed InfoGAN (DPD-InfoGAN)}
\label{DPInfoGAN2}
\end{algorithm}

\section{Experiments}
We ran experiments on the MNIST dataset to analyze the trade-off between privacy and the quality and semantics of generated images. The batch size was set to 64, the number of epochs to 50, and $\delta$ to $10^{-5}$.  We fix the learning rate for the Adam optimizer to $0.0002$, and sample two continuous codes from a uniform distribution between $[-1,1]$. We use three fractionally-strided convolutions for the generator and three convolutions for the discriminator. The Q network consists of four convolutional layers. Batch normalization is applied in all the layers. LeakyRELU is used in discriminator and Q network, while the generator uses RELU activation.

\begin{figure}[h]\centering
\subfloat[InfoGAN]{\label{2a}\includegraphics[width=.48\linewidth]{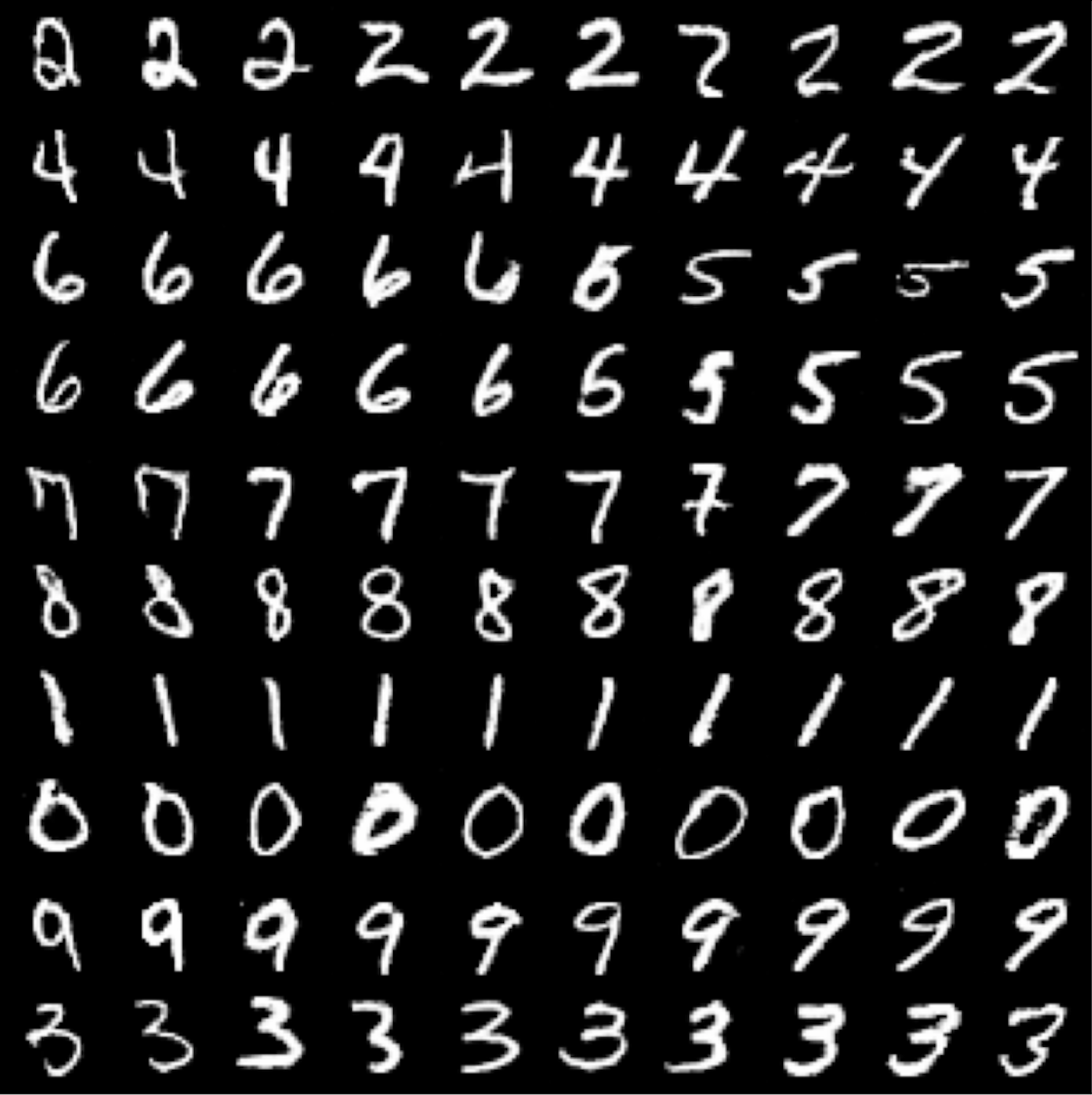}}\hfill
\subfloat[DP-InfoGAN ($\epsilon=1$)]{\label{2b}\includegraphics[width=.48\linewidth]{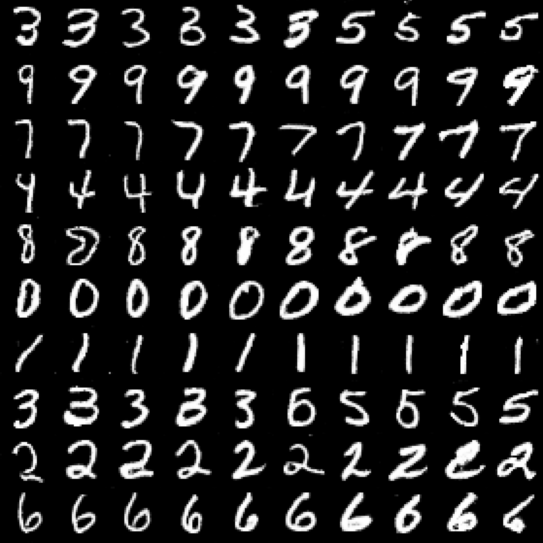}}\par
\subfloat[DP-InfoGAN ($\epsilon=0.1$)]{\label{2c}\includegraphics[width=.48\linewidth]{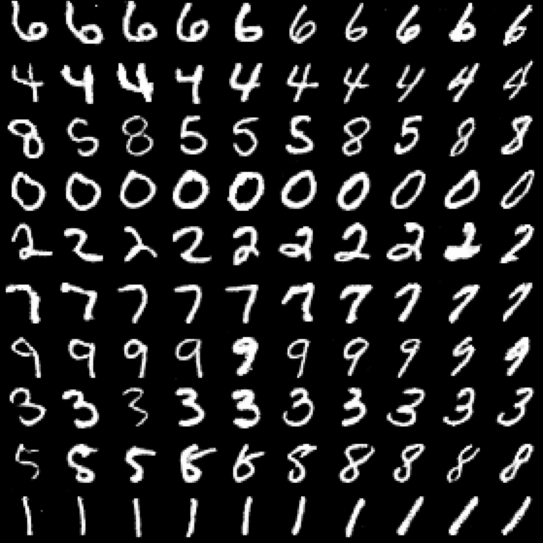}}
\caption{Rotation of digits}
\label{continuos}
\end{figure}

\begin{figure}[h]
\begin{subfigure}{.23\textwidth}
\centering
\includegraphics[width=\linewidth]{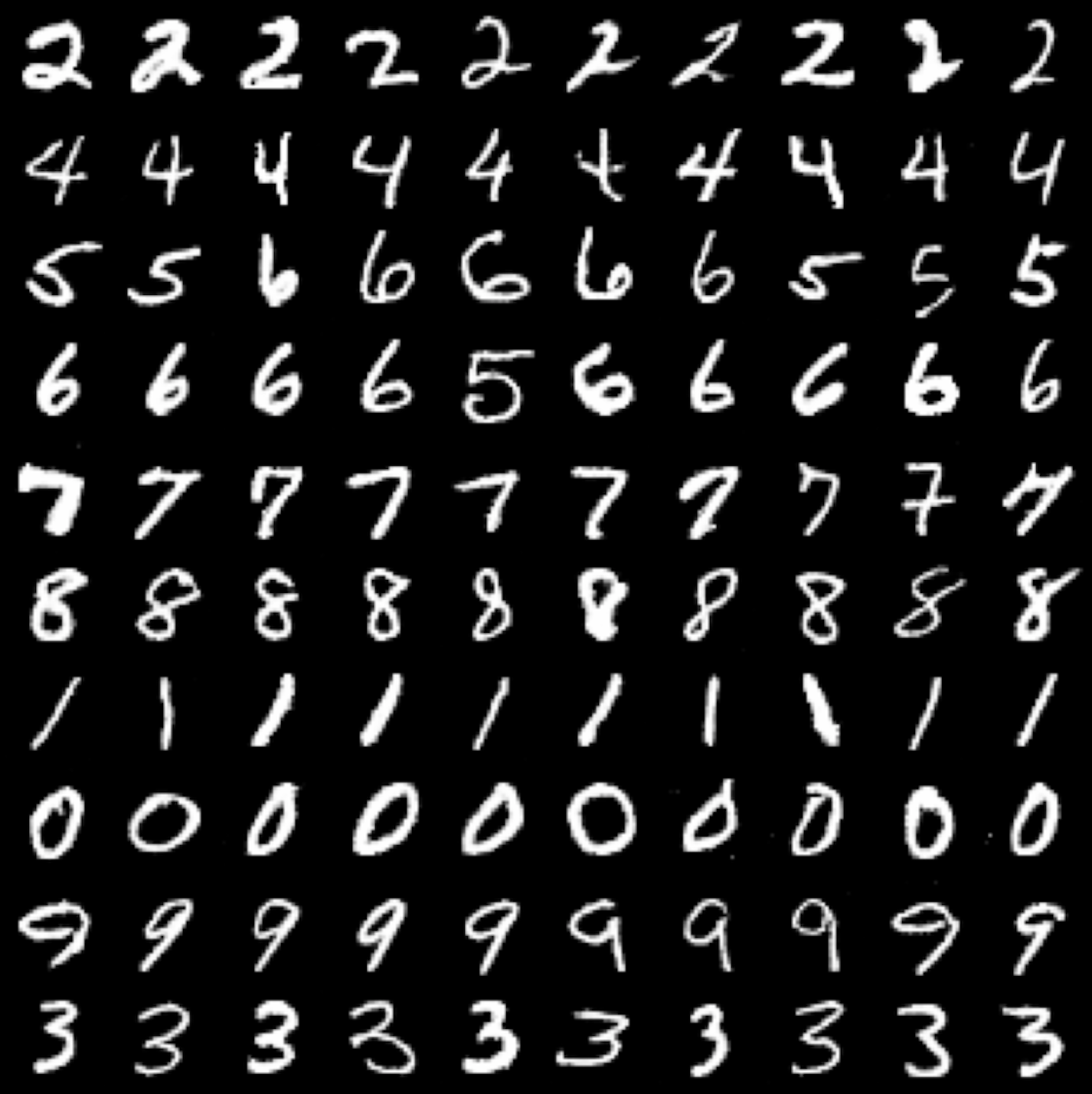}
\caption{InfoGAN}
\label{3a}

\end{subfigure}\hfill
\begin{subfigure}{.23\textwidth}
\centering
\includegraphics[width=\linewidth]{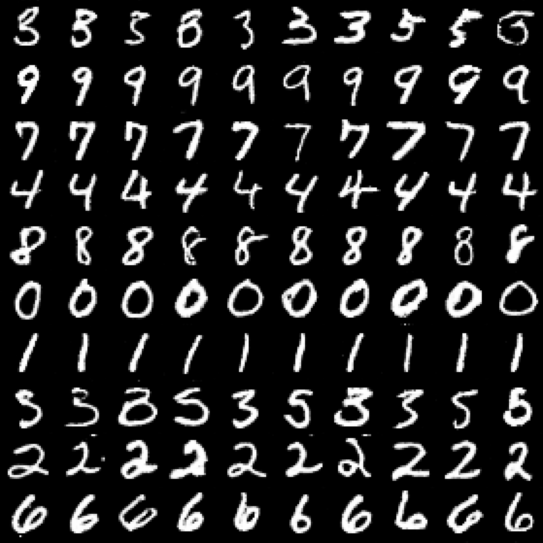}
\caption{DP-InfoGAN ($\epsilon=1$)}
\label{3b}
\end{subfigure}

\caption{Identity of Numbers}
\label{discrete}

\end{figure}

We mainly employ qualitative evaluation as approaches such as Inception Score \cite{salimans1606improved} and Frechnet Inception Distance \cite{heusel2017gans} do not reflect the quality of rotation or thickness factors. First, we compare the results obtained from InfoGAN and DP-InfoGAN on a single client model. In Figure \ref{discrete}, we see that with privacy guarantees, the model has trouble differentiating between close digits (such as 3 and 5 in Figure \ref{3b}) but still is able to generate high-quality images. In addition, as shown in Figure \ref{2b}, DP-InfoGAN faces minor issues disentangling the latent space (i.e.) the results display changes in both thickness and rotation while we try to preserve only rotation of digits. Therefore, DP-InfoGAN preserves the quality of the images to an extent, but starts losing control over the attributes of images. For smaller values of $\epsilon$, it becomes harder to facilitate this control. We see that in Figure \ref{2c}, the thickness and rotation of the digits get more entangled when compared to Figures \ref{2a}, \ref{2b}.

\begin{figure}[h]
\begin{subfigure}{.23\textwidth}
\centering
\includegraphics[width=\linewidth]{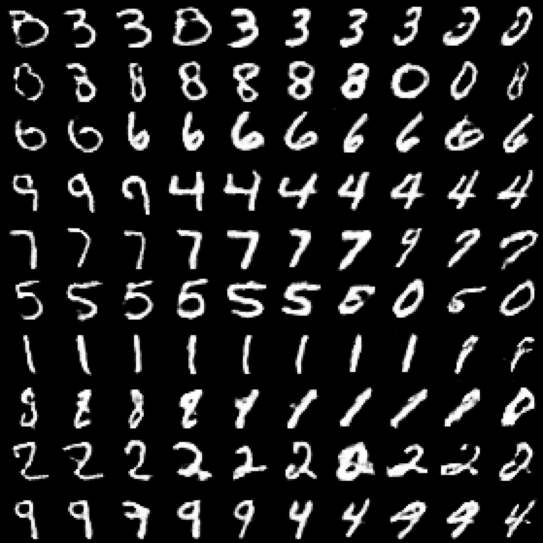}
\caption{Number of Clients : 1}
\label{4a}
\end{subfigure}\hfill
\begin{subfigure}{.23\textwidth}
\centering
\includegraphics[width=\linewidth]{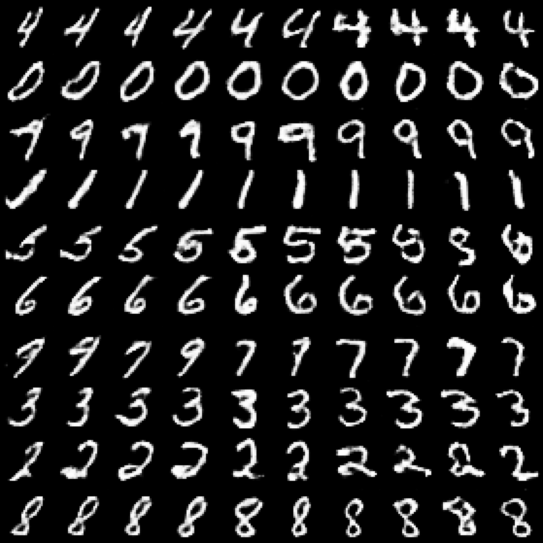}
\caption{Number of Clients : 10}
\label{4b}
\end{subfigure}\hfill

\caption{Rotation of digits ($\epsilon=10$)}
\label{}

\end{figure}

\begin{figure}[h]
\begin{subfigure}{.23\textwidth}
\centering
\includegraphics[width=\linewidth]{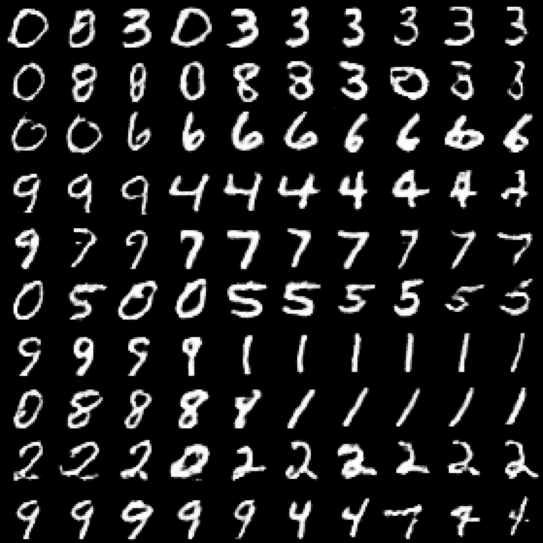}
\caption{Number of Clients : 1}
\label{5a}
\end{subfigure}\hfill
\begin{subfigure}{.23\textwidth}
\centering
\includegraphics[width=\linewidth]{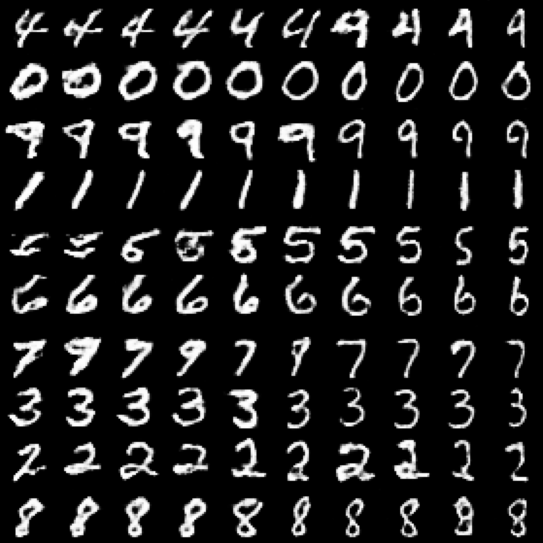}
\caption{Number of Clients : 10}
\label{5b}
\end{subfigure}

\caption{Thickness of digits ($\epsilon=10$)}
\label{}

\end{figure}

In the distributed setting (DPD-InfoGAN), we use the same configuration and simulate experiments with each client having a subset of the MNIST data (non-overlapping). To validate our algorithm and prove that a shared Q-network can capture all possible variances in images, we run experiments with 10 clients (each having 6000 images) in a distributed setting and compare it with a single client having 6000 images. We observe that, in the distributed setting, the generated images display a varied change in thickness and rotation when compared to the non-distributed setting. In Figures \ref{4b}, \ref{5b}, we see more variety in rotation and thickness when compared to a single client setting as shown in Figures \ref{4a}, \ref{5a}. For example, digits 1 and 8 in Figure \ref{4b} have more variations in rotations than in Figure \ref{4a}. This indicates that even when a client does not have variations in its training images, it can still generate those variations as the shared Q-network is continuously updated on the clients' datasets in a privacy-preserving manner and captures all possible feature variances present in the datasets of other clients.

\begin{figure}[h]\centering
\subfloat[DP-InfoGAN($\epsilon=10$)]{\label{6a}\includegraphics[width=.48\linewidth]{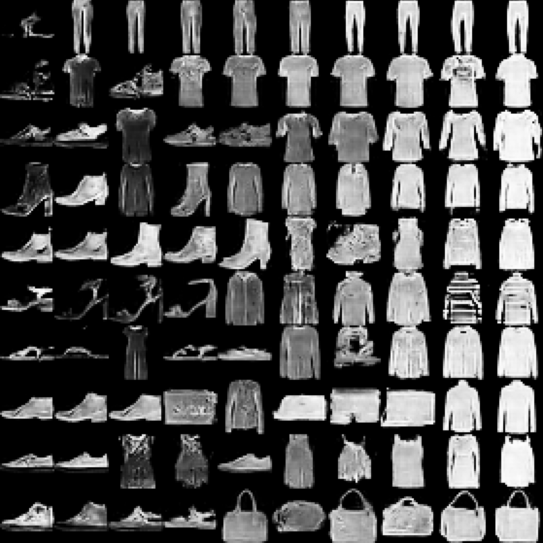}}\hfill
\subfloat[DP-InfoGAN ($\epsilon=1$)]{\label{6b}\includegraphics[width=.48\linewidth]{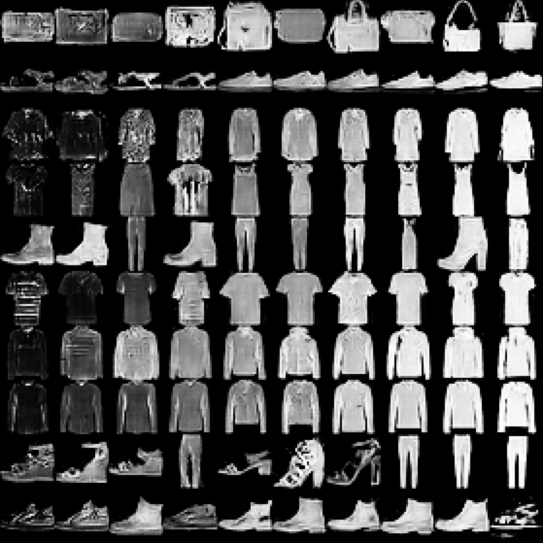}}\par
\subfloat[DP-InfoGAN ($\epsilon=0.1$)]{\label{6c}\includegraphics[width=.48\linewidth]{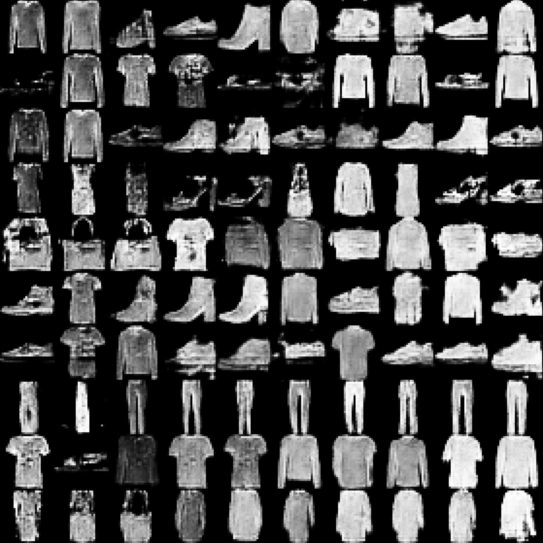}}
\caption{Variation in thickness of images for varying $\epsilon$}
\label{6}
\end{figure}

We ran experiments on the FashionMNIST dataset using the same setup as used for the MNIST dataset. Since the data points in the FashionMNIST dataset do not vary for rotation, we demonstrate the results for the thickness factor. In the non-distributed setting (DP-InfoGAN), we again find that as the value of epsilon decreases (more privacy), the model has trouble differentiating between objects like shoes and shirts. The amount of thickness variation also reduces as the privacy guarantees increase, as shown in Figure \ref{6}.

\begin{figure}[h]
\begin{subfigure}{.23\textwidth}
\centering
\includegraphics[width=\linewidth]{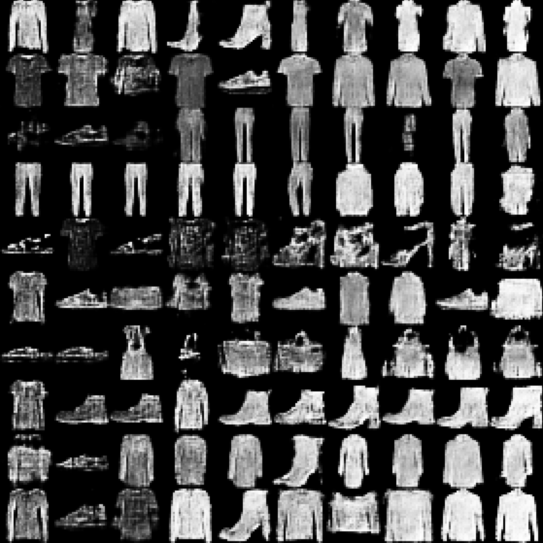}
\caption{Number of Clients : 1}
\label{7a}
\end{subfigure}\hfill
\begin{subfigure}{.23\textwidth}
\centering
\includegraphics[width=\linewidth]{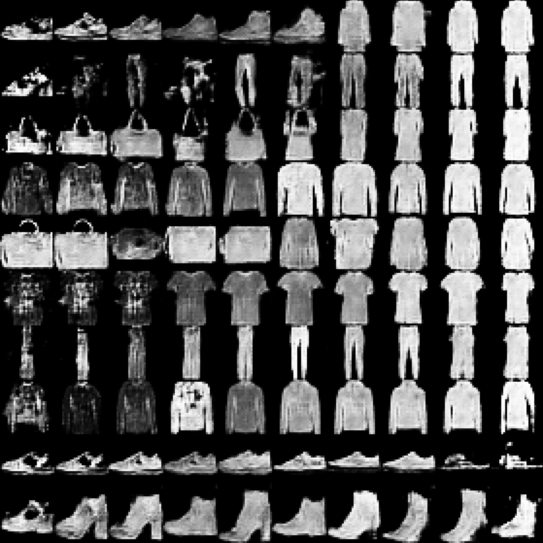}
\caption{Number of Clients : 10}
\label{7b}
\end{subfigure}

\caption{Variation in thickness of images for varying number of clients ($\epsilon=0.1$)}
\label{}

\end{figure}

In the distributed setting (Figures \ref{7a}, \ref{7b}), we again find that with an increase in the number of clients, the shared Q-network helps in learning more variations for the thickness factor, when compared to a single client with the same number of images.

\section{Conclusion and Future Work}
In this paper, we propose a privacy-preserving version of InfoGAN (DP-InfoGAN) that guarantees the privacy of the training data samples. Our results show that DP-InfoGAN can synthesize high-quality images with control on image attributes. Our framework can keep track of the privacy budget spent by using Moment Accountant or Renyi DP accountant. We also extend the framework to a distributed setting (DPD-InfoGAN) by using a shared Q network. We show that our training paradigm in the distributed setting captures varied image characteristics even when each client has limited data. As part of future work, we plan to explore privacy-preserving and distributed/federated versions of CycleGANs, BigGANs, etc.

\bibliographystyle{ACM-Reference-Format}
\bibliography{sample-base}


\begin{thebibliography}{24}


\ifx \showCODEN    \undefined \def \showCODEN     #1{\unskip}     \fi
\ifx \showDOI      \undefined \def \showDOI       #1{#1}\fi
\ifx \showISBNx    \undefined \def \showISBNx     #1{\unskip}     \fi
\ifx \showISBNxiii \undefined \def \showISBNxiii  #1{\unskip}     \fi
\ifx \showISSN     \undefined \def \showISSN      #1{\unskip}     \fi
\ifx \showLCCN     \undefined \def \showLCCN      #1{\unskip}     \fi
\ifx \shownote     \undefined \def \shownote      #1{#1}          \fi
\ifx \showarticletitle \undefined \def \showarticletitle #1{#1}   \fi
\ifx \showURL      \undefined \def \showURL       {\relax}        \fi
\providecommand\bibfield[2]{#2}
\providecommand\bibinfo[2]{#2}
\providecommand\natexlab[1]{#1}
\providecommand\showeprint[2][]{arXiv:#2}

\bibitem[\protect\citeauthoryear{Abadi, Chu, Goodfellow, McMahan, Mironov,
  Talwar, and Zhang}{Abadi et~al\mbox{.}}{2016}]%
        {abadi2016deep}
\bibfield{author}{\bibinfo{person}{Martin Abadi}, \bibinfo{person}{Andy Chu},
  \bibinfo{person}{Ian Goodfellow}, \bibinfo{person}{H~Brendan McMahan},
  \bibinfo{person}{Ilya Mironov}, \bibinfo{person}{Kunal Talwar}, {and}
  \bibinfo{person}{Li Zhang}.} \bibinfo{year}{2016}\natexlab{}.
\newblock \showarticletitle{Deep learning with differential privacy}. In
  \bibinfo{booktitle}{\emph{Proceedings of the 2016 ACM SIGSAC Conference on
  Computer and Communications Security}}. \bibinfo{pages}{308--318}.
\newblock


\bibitem[\protect\citeauthoryear{Chen, Xiang, Xue, Li, Borisov, Kaarfar, and
  Zhu}{Chen et~al\mbox{.}}{2018}]%
        {chen2018differentially}
\bibfield{author}{\bibinfo{person}{Qingrong Chen}, \bibinfo{person}{Chong
  Xiang}, \bibinfo{person}{Minhui Xue}, \bibinfo{person}{Bo Li},
  \bibinfo{person}{Nikita Borisov}, \bibinfo{person}{Dali Kaarfar}, {and}
  \bibinfo{person}{Haojin Zhu}.} \bibinfo{year}{2018}\natexlab{}.
\newblock \showarticletitle{Differentially private data generative models}.
\newblock \bibinfo{journal}{\emph{arXiv preprint arXiv:1812.02274}}
  (\bibinfo{year}{2018}).
\newblock


\bibitem[\protect\citeauthoryear{Chen, Duan, Houthooft, Schulman, Sutskever,
  and Abbeel}{Chen et~al\mbox{.}}{2016}]%
        {chen2016infogan}
\bibfield{author}{\bibinfo{person}{Xi Chen}, \bibinfo{person}{Yan Duan},
  \bibinfo{person}{Rein Houthooft}, \bibinfo{person}{John Schulman},
  \bibinfo{person}{Ilya Sutskever}, {and} \bibinfo{person}{Pieter Abbeel}.}
  \bibinfo{year}{2016}\natexlab{}.
\newblock \showarticletitle{Infogan: Interpretable representation learning by
  information maximizing generative adversarial nets}. In
  \bibinfo{booktitle}{\emph{Advances in neural information processing
  systems}}. \bibinfo{pages}{2172--2180}.
\newblock


\bibitem[\protect\citeauthoryear{Dwork, McSherry, Nissim, and Smith}{Dwork
  et~al\mbox{.}}{2006}]%
        {dwork2006calibrating}
\bibfield{author}{\bibinfo{person}{Cynthia Dwork}, \bibinfo{person}{Frank
  McSherry}, \bibinfo{person}{Kobbi Nissim}, {and} \bibinfo{person}{Adam
  Smith}.} \bibinfo{year}{2006}\natexlab{}.
\newblock \showarticletitle{Calibrating noise to sensitivity in private data
  analysis}. In \bibinfo{booktitle}{\emph{Theory of cryptography conference}}.
  Springer, \bibinfo{pages}{265--284}.
\newblock


\bibitem[\protect\citeauthoryear{Dwork, Roth, et~al\mbox{.}}{Dwork
  et~al\mbox{.}}{2014}]%
        {dwork2014algorithmic}
\bibfield{author}{\bibinfo{person}{Cynthia Dwork}, \bibinfo{person}{Aaron
  Roth}, {et~al\mbox{.}}} \bibinfo{year}{2014}\natexlab{}.
\newblock \showarticletitle{The algorithmic foundations of differential
  privacy.}
\newblock \bibinfo{journal}{\emph{Foundations and Trends in Theoretical
  Computer Science}} \bibinfo{volume}{9}, \bibinfo{number}{3-4}
  (\bibinfo{year}{2014}), \bibinfo{pages}{211--407}.
\newblock


\bibitem[\protect\citeauthoryear{Frigerio, de~Oliveira, Gomez, and
  Duverger}{Frigerio et~al\mbox{.}}{2019}]%
        {frigerio2019differentially}
\bibfield{author}{\bibinfo{person}{Lorenzo Frigerio},
  \bibinfo{person}{Anderson~Santana de Oliveira}, \bibinfo{person}{Laurent
  Gomez}, {and} \bibinfo{person}{Patrick Duverger}.}
  \bibinfo{year}{2019}\natexlab{}.
\newblock \showarticletitle{Differentially private generative adversarial
  networks for time series, continuous, and discrete open data}. In
  \bibinfo{booktitle}{\emph{IFIP International Conference on ICT Systems
  Security and Privacy Protection}}. Springer, \bibinfo{pages}{151--164}.
\newblock


\bibitem[\protect\citeauthoryear{Goodfellow, Pouget-Abadie, Mirza, Xu,
  Warde-Farley, Ozair, Courville, and Bengio}{Goodfellow et~al\mbox{.}}{2014}]%
        {goodfellow2014generative}
\bibfield{author}{\bibinfo{person}{Ian Goodfellow}, \bibinfo{person}{Jean
  Pouget-Abadie}, \bibinfo{person}{Mehdi Mirza}, \bibinfo{person}{Bing Xu},
  \bibinfo{person}{David Warde-Farley}, \bibinfo{person}{Sherjil Ozair},
  \bibinfo{person}{Aaron Courville}, {and} \bibinfo{person}{Yoshua Bengio}.}
  \bibinfo{year}{2014}\natexlab{}.
\newblock \showarticletitle{Generative adversarial nets}. In
  \bibinfo{booktitle}{\emph{Advances in neural information processing
  systems}}. \bibinfo{pages}{2672--2680}.
\newblock


\bibitem[\protect\citeauthoryear{Hayes, Melis, Danezis, and
  De~Cristofaro}{Hayes et~al\mbox{.}}{2019}]%
        {hayes2019logan}
\bibfield{author}{\bibinfo{person}{Jamie Hayes}, \bibinfo{person}{Luca Melis},
  \bibinfo{person}{George Danezis}, {and} \bibinfo{person}{Emiliano
  De~Cristofaro}.} \bibinfo{year}{2019}\natexlab{}.
\newblock \showarticletitle{LOGAN: Membership inference attacks against
  generative models}.
\newblock \bibinfo{journal}{\emph{Proceedings on Privacy Enhancing
  Technologies}} \bibinfo{volume}{2019}, \bibinfo{number}{1}
  (\bibinfo{year}{2019}), \bibinfo{pages}{133--152}.
\newblock


\bibitem[\protect\citeauthoryear{Heusel, Ramsauer, Unterthiner, Nessler, and
  Hochreiter}{Heusel et~al\mbox{.}}{2017}]%
        {heusel2017gans}
\bibfield{author}{\bibinfo{person}{Martin Heusel}, \bibinfo{person}{Hubert
  Ramsauer}, \bibinfo{person}{Thomas Unterthiner}, \bibinfo{person}{Bernhard
  Nessler}, {and} \bibinfo{person}{Sepp Hochreiter}.}
  \bibinfo{year}{2017}\natexlab{}.
\newblock \showarticletitle{Gans trained by a two time-scale update rule
  converge to a local nash equilibrium}. In \bibinfo{booktitle}{\emph{Advances
  in neural information processing systems}}. \bibinfo{pages}{6626--6637}.
\newblock


\bibitem[\protect\citeauthoryear{Jordon, Yoon, and van~der Schaar}{Jordon
  et~al\mbox{.}}{2018}]%
        {jordon2018pate}
\bibfield{author}{\bibinfo{person}{James Jordon}, \bibinfo{person}{Jinsung
  Yoon}, {and} \bibinfo{person}{Mihaela van~der Schaar}.}
  \bibinfo{year}{2018}\natexlab{}.
\newblock \showarticletitle{PATE-GAN: Generating synthetic data with
  differential privacy guarantees}. In \bibinfo{booktitle}{\emph{International
  Conference on Learning Representations}}.
\newblock


\bibitem[\protect\citeauthoryear{Kingma and Ba}{Kingma and Ba}{2014}]%
        {kingma2014adam}
\bibfield{author}{\bibinfo{person}{Diederik~P Kingma} {and}
  \bibinfo{person}{Jimmy Ba}.} \bibinfo{year}{2014}\natexlab{}.
\newblock \showarticletitle{Adam: A method for stochastic optimization}.
\newblock \bibinfo{journal}{\emph{arXiv preprint arXiv:1412.6980}}
  (\bibinfo{year}{2014}).
\newblock


\bibitem[\protect\citeauthoryear{Kone{\v{c}}n{\`y}, McMahan, Yu, Richt{\'a}rik,
  Suresh, and Bacon}{Kone{\v{c}}n{\`y} et~al\mbox{.}}{2016}]%
        {konevcny2016federated}
\bibfield{author}{\bibinfo{person}{Jakub Kone{\v{c}}n{\`y}},
  \bibinfo{person}{H~Brendan McMahan}, \bibinfo{person}{Felix~X Yu},
  \bibinfo{person}{Peter Richt{\'a}rik}, \bibinfo{person}{Ananda~Theertha
  Suresh}, {and} \bibinfo{person}{Dave Bacon}.}
  \bibinfo{year}{2016}\natexlab{}.
\newblock \showarticletitle{Federated learning: Strategies for improving
  communication efficiency}.
\newblock \bibinfo{journal}{\emph{arXiv preprint arXiv:1610.05492}}
  (\bibinfo{year}{2016}).
\newblock


\bibitem[\protect\citeauthoryear{LeCun, Cortes, and Burges}{LeCun
  et~al\mbox{.}}{2010}]%
        {lecun2010mnist}
\bibfield{author}{\bibinfo{person}{Yann LeCun}, \bibinfo{person}{Corinna
  Cortes}, {and} \bibinfo{person}{CJ Burges}.} \bibinfo{year}{2010}\natexlab{}.
\newblock \showarticletitle{MNIST handwritten digit database}.
\newblock \bibinfo{journal}{\emph{ATT Labs [Online]. Available:
  http://yann.lecun.com/exdb/mnist}}  \bibinfo{volume}{2}
  (\bibinfo{year}{2010}).
\newblock


\bibitem[\protect\citeauthoryear{Li, Swersky, and Zemel}{Li
  et~al\mbox{.}}{2015}]%
        {li2015generative}
\bibfield{author}{\bibinfo{person}{Yujia Li}, \bibinfo{person}{Kevin Swersky},
  {and} \bibinfo{person}{Rich Zemel}.} \bibinfo{year}{2015}\natexlab{}.
\newblock \showarticletitle{Generative moment matching networks}. In
  \bibinfo{booktitle}{\emph{International Conference on Machine Learning}}.
  \bibinfo{pages}{1718--1727}.
\newblock


\bibitem[\protect\citeauthoryear{Makhzani, Shlens, Jaitly, Goodfellow, and
  Frey}{Makhzani et~al\mbox{.}}{2015}]%
        {makhzani2015adversarial}
\bibfield{author}{\bibinfo{person}{Alireza Makhzani}, \bibinfo{person}{Jonathon
  Shlens}, \bibinfo{person}{Navdeep Jaitly}, \bibinfo{person}{Ian Goodfellow},
  {and} \bibinfo{person}{Brendan Frey}.} \bibinfo{year}{2015}\natexlab{}.
\newblock \showarticletitle{Adversarial autoencoders}.
\newblock \bibinfo{journal}{\emph{arXiv preprint arXiv:1511.05644}}
  (\bibinfo{year}{2015}).
\newblock


\bibitem[\protect\citeauthoryear{Melis, Song, De~Cristofaro, and
  Shmatikov}{Melis et~al\mbox{.}}{2019}]%
        {melis2019exploiting}
\bibfield{author}{\bibinfo{person}{Luca Melis}, \bibinfo{person}{Congzheng
  Song}, \bibinfo{person}{Emiliano De~Cristofaro}, {and}
  \bibinfo{person}{Vitaly Shmatikov}.} \bibinfo{year}{2019}\natexlab{}.
\newblock \showarticletitle{Exploiting unintended feature leakage in
  collaborative learning}. In \bibinfo{booktitle}{\emph{2019 IEEE Symposium on
  Security and Privacy (SP)}}. IEEE, \bibinfo{pages}{691--706}.
\newblock


\bibitem[\protect\citeauthoryear{Mironov}{Mironov}{2017}]%
        {mironov2017renyi}
\bibfield{author}{\bibinfo{person}{Ilya Mironov}.}
  \bibinfo{year}{2017}\natexlab{}.
\newblock \showarticletitle{R{\'e}nyi differential privacy}. In
  \bibinfo{booktitle}{\emph{2017 IEEE 30th Computer Security Foundations
  Symposium (CSF)}}. IEEE, \bibinfo{pages}{263--275}.
\newblock


\bibitem[\protect\citeauthoryear{Nasr, Shokri, and Houmansadr}{Nasr
  et~al\mbox{.}}{2018}]%
        {nasr2018comprehensive}
\bibfield{author}{\bibinfo{person}{Milad Nasr}, \bibinfo{person}{Reza Shokri},
  {and} \bibinfo{person}{Amir Houmansadr}.} \bibinfo{year}{2018}\natexlab{}.
\newblock \showarticletitle{Comprehensive privacy analysis of deep learning:
  Stand-alone and federated learning under passive and active white-box
  inference attacks}.
\newblock \bibinfo{journal}{\emph{arXiv preprint arXiv:1812.00910}}
  (\bibinfo{year}{2018}).
\newblock


\bibitem[\protect\citeauthoryear{Rezende, Mohamed, and Wierstra}{Rezende
  et~al\mbox{.}}{2014}]%
        {rezende2014stochastic}
\bibfield{author}{\bibinfo{person}{Danilo~Jimenez Rezende},
  \bibinfo{person}{Shakir Mohamed}, {and} \bibinfo{person}{Daan Wierstra}.}
  \bibinfo{year}{2014}\natexlab{}.
\newblock \showarticletitle{Stochastic backpropagation and approximate
  inference in deep generative models}.
\newblock \bibinfo{journal}{\emph{arXiv preprint arXiv:1401.4082}}
  (\bibinfo{year}{2014}).
\newblock


\bibitem[\protect\citeauthoryear{Salimans, Goodfellow, Zaremba, Cheung,
  Radford, and Chen}{Salimans et~al\mbox{.}}{[n.d.]}]%
        {salimans1606improved}
\bibfield{author}{\bibinfo{person}{Tim Salimans}, \bibinfo{person}{Ian
  Goodfellow}, \bibinfo{person}{Wojciech Zaremba}, \bibinfo{person}{Vicki
  Cheung}, \bibinfo{person}{Alec Radford}, {and} \bibinfo{person}{Xi Chen}.}
  \bibinfo{year}{[n.d.]}\natexlab{}.
\newblock \showarticletitle{Improved techniques for training Gans. arXiv 2016}.
\newblock \bibinfo{journal}{\emph{arXiv preprint arXiv:1606.03498}}
  (\bibinfo{year}{[n.\,d.]}).
\newblock


\bibitem[\protect\citeauthoryear{Shokri, Stronati, Song, and Shmatikov}{Shokri
  et~al\mbox{.}}{2017}]%
        {shokri2017membership}
\bibfield{author}{\bibinfo{person}{Reza Shokri}, \bibinfo{person}{Marco
  Stronati}, \bibinfo{person}{Congzheng Song}, {and} \bibinfo{person}{Vitaly
  Shmatikov}.} \bibinfo{year}{2017}\natexlab{}.
\newblock \showarticletitle{Membership inference attacks against machine
  learning models}. In \bibinfo{booktitle}{\emph{2017 IEEE Symposium on
  Security and Privacy (SP)}}. IEEE, \bibinfo{pages}{3--18}.
\newblock


\bibitem[\protect\citeauthoryear{Torkzadehmahani, Kairouz, and
  Paten}{Torkzadehmahani et~al\mbox{.}}{2019}]%
        {torkzadehmahani2019dp}
\bibfield{author}{\bibinfo{person}{Reihaneh Torkzadehmahani},
  \bibinfo{person}{Peter Kairouz}, {and} \bibinfo{person}{Benedict Paten}.}
  \bibinfo{year}{2019}\natexlab{}.
\newblock \showarticletitle{Dp-cgan: Differentially private synthetic data and
  label generation}. In \bibinfo{booktitle}{\emph{Proceedings of the IEEE
  Conference on Computer Vision and Pattern Recognition Workshops}}.
  \bibinfo{pages}{0--0}.
\newblock


\bibitem[\protect\citeauthoryear{Voigt and Von~dem Bussche}{Voigt and Von~dem
  Bussche}{2017}]%
        {voigt2017eu}
\bibfield{author}{\bibinfo{person}{Paul Voigt} {and} \bibinfo{person}{Axel
  Von~dem Bussche}.} \bibinfo{year}{2017}\natexlab{}.
\newblock \showarticletitle{The eu general data protection regulation (gdpr)}.
\newblock \bibinfo{journal}{\emph{A Practical Guide, 1st Ed., Cham: Springer
  International Publishing}} (\bibinfo{year}{2017}).
\newblock


\bibitem[\protect\citeauthoryear{Xie, Lin, Wang, Wang, and Zhou}{Xie
  et~al\mbox{.}}{2018}]%
        {xie2018differentially}
\bibfield{author}{\bibinfo{person}{Liyang Xie}, \bibinfo{person}{Kaixiang Lin},
  \bibinfo{person}{Shu Wang}, \bibinfo{person}{Fei Wang}, {and}
  \bibinfo{person}{Jiayu Zhou}.} \bibinfo{year}{2018}\natexlab{}.
\newblock \showarticletitle{Differentially private generative adversarial
  network}.
\newblock \bibinfo{journal}{\emph{arXiv preprint arXiv:1802.06739}}
  (\bibinfo{year}{2018}).
\newblock


\end{thebibliography}


\end{document}